\documentclass[12pt]{article}
\usepackage{graphicx}
\usepackage{subfigure}
\usepackage{amsmath,amsthm}
\usepackage{amssymb,amsfonts}
\usepackage{authblk}
\usepackage{url}
\usepackage{booktabs}

\newcommand{\tabincell}[2]{\begin{tabular}{@{}#1@{}}#2\end{tabular}}

\newtheorem{definition}{\textbf{Definition}}

\newtheorem{theorem}{\textbf{Theorem}}

\newtheorem{corollary}{\textbf{Corollary}}

\title{Variable Selection with Copula Entropy}
\author{Jian MA\thanks{Email: majian@hitachi.cn}}
\affil{\normalsize Hitachi (China) Research \& Development Corporation}

\begin{document}

\maketitle

\begin{abstract}
	\noindent
Variable selection is of significant importance for classification and regression tasks in machine learning and statistical applications where both predictability and explainability are needed. In this paper, a Copula Entropy (CE) based method for variable selection which use CE based ranks to select variables is proposed. The method is both model-free and tuning-free. Comparison experiments between the proposed method and traditional variable selection methods, such as Distance Correlation, Hilbert-Schmidt Independence Criterion, Stepwise Selection, regularized generalized linear models and Adaptive LASSO, were conducted on the UCI heart disease data. Experimental results show that CE based method can select the `right' variables out more effectively and derive better interpretable results than traditional methods do without sacrificing accuracy performance. It is believed that CE based variable selection can help to build more explainable models.
\end{abstract}

\section{Introduction}
\label{s:introduction}
\subsection{Variable Selection Problem}
Variable selection is one of the old and widely studied model selection problems in statistics and machine learning \cite{guyon2003,george2000}. The problem arises when one wants to model the relationship between a variable of interest (response) and a (large) amount of potential explainable variables (predictors) but only a subset of latter explainable variables may be relevant to the former variable. The aim of the problem is to select a subset of variables under certain criteria. The criteria of selection are variable’s ability of prediction and interpretation.

\subsection{Existing Methods}
\subsubsection{Measure based Variable Selection}
\label{subsec:measure}
The most natural way of variable selection is based on statistical association measures between response and individual predictor. Due to its simplicity and interpretability, measure based selection enjoys widely adoption and empirical successes in practice. The most common traditional measure is Pearson Correlation Coefficient (CC) for linear models. However, the application of CC assumes Gassianity, which is unrealistic for most non-linear and non-Gaussian cases. For nonlinear dependence measure, some may considered Mutual Information (MI) in information theory \cite{15}, but only a few applications are reported due to notorious difficulty of estimating MI. Besides MI, several other nonlinear dependence measures were proposed.

Distance Correlation (dCor) is a nonlinear generalization of traditional correlation concept proposed by  Sz\'ekely, et al \cite{dcor1, dcor2}. It generalizes bivariate second-order correlation to multivariate nonlinear cases via distance covariance. dCor between random vectors $X$ and $Y$ is defined as
\begin{equation}
\mathbf{dCor}(X,Y) = \frac{\nu^2(X,Y)}{\sqrt{\nu^2(X)\nu^2(Y)}},
\end{equation}
where $\nu^2(X,Y)$ is distance covariance defined with characteristic function $f$ as
\begin{equation}
\nu^2(X,Y;w) = \|f_{X,Y}(t,s) - f_X(t)f_Y(s)\|_w^2.
\end{equation}
Here, $\|\cdot\|_w$ is the norm in the weighted $L_2$ function space defined with positive weight function $w(\cdot,\cdot)$ \cite{dcor1,dcor2}. dCor characterizes independence: $\mathbf{dCor}(X,Y) \geq 0$, and $\mathbf{dCor}(X,Y) = 0$ if and only if $X,Y$ are independent. dCor has been proposed as a tool for variable selection \cite{dcorvar}.

Hilbert-Schmidt Independence Criterion (HSIC) is another widely studied independence measure \cite{hsic2} and it has multivariate version -- d-variable HSIC (dHSIC) \cite{hsic}. dHSIC defines a nonlinear dependence measure in Reproducing Kernel Hilbert Spaces (RKHS) with kernel function, as follows:
\begin{equation}
\mathbf{dHSIC}(P(X_1,\cdots,X_d)) = \|\Pi(P(X_1)\otimes,\cdots,\otimes P(X_d))-\Pi(P(X_1,\cdots,X_d))\|,
\end{equation}
where $\Pi$ is kernel mean embedding function, and $\otimes$ is tensor products of kernels. dHSIC can be considered as the distance in RKHS between the embeddings of joint distribution and margins. dHSIC also characterizes independence: $\mathbf{dHSIC}(P(X_1,\ldots,X_d))=0$ if and only if $X_1,\ldots, X_d$ are independent so it is natural to apply it to variable selection problem \cite{dhsicvar}. 

\subsubsection{Stepwise Selection with Information Criteria}
Stepwise Selection is a standard approach for variable selection, usually on linear regression models, which sequentially selecting or eliminating the predictors once at a time based on certain criteria. Two of the main criteria are Akaike Information Criterion (AIC) \cite{aic} and Bayesian Information Criterion (BIC) \cite{bic}. Let $l$ denotes the log maximum likelihood of the model, $p$  denotes the number of free parameters of the model and $N$ denotes the number of observation, and then AIC is defined as
\begin{equation}\label{eq:aic}
\mathbf{AIC} = -2l + 2p,
\end{equation}
and BIC is defined as
\begin{equation}\label{eq:bic}
\mathbf{BIC} = -2l+p\log N.
\end{equation}
It can be learned from the definitions that both criteria are defined as penalized likelihood criteria that try to achieve a balance between goodness of fit (likelihood) and penalty on overfitting (number of free parameters) for model selection problem.

\subsubsection{Regularized Generalized Linear Models}
Linear Regression (LR) is the commonly used model in most of the researches that study the relationships under the assumption of linearity. Generalized Linear Models (GLMs) are a group of variants of LR, such as Logistic Regression and Poisson Regression, which introduce nonlinear response by means of link function. Due to their poor ability of variable selection, LR or GLMs are not applied directed to many cases for model selection, especially to high dimensional problems. To tackle this issue, regularization techniques are introduced to formalize a new learning problem from GLM problems under the assumption of sparsity. Let $y$ denotes response, $\mathbf{X}$ denote predictors, and $\beta$ denote coefficients to be estimated, and then regularized GLMs solve the following problem:
\begin{equation}\label{eq:rglm}
\min_{\beta}{\{L(\beta;y,\mathbf{X})+\lambda_1 \|\beta\|_1 +\lambda_2 \|\beta\|_2^2 \}},
\end{equation}
where $L(\cdot)$ denotes likelihood function, $\|\beta\|_i$ denote $i$th norm of coefficients $\beta$, and $\lambda_i$ denote tuning parameters.

Three main variants of this regularized problem are defined by  tuning the parameters $\lambda_1,\lambda_2$. The problem for the case where $\lambda_1 =0,\lambda_2 >0$ is called Ridge Regression \cite{ridge}. The Least Absolute Shrinkage and Selection Operator (LASSO) problem corresponds to the cases where $\lambda_1 >0,\lambda_2 =0$ \cite{lasso}. The problem for $\lambda_1,\lambda_2 >0$ is called Elastic Net \cite{elasticnet}. For more variants of regularized GLMs, please refer to \cite{fan2010,wu2014} and references therein. 

Regularized GLMs select a subset of variable by means of shrinking the non-zero coefficients of variables. Suppose the `true' model has a group of sparse coefficients, an estimator is said to has oracle property if it can estimate this coefficients effectively and asymptotically \cite{fan2001oracle}. Zou \cite{zou2006ada} has shown that the LASSO estimator has no oracle property. To address this issue, he proposed the Adaptive LASSO with the adaptive weights technique and demontrated its oracle property under regularity condtions \cite{zou2006ada}.

\subsection{Limitations of the Existing Methods}
When using Stepwise Selection with different criteria, one may confuse which criterion is fitful to the given problem since different criteria derive different learning models. Comparison between the definition of AIC \eqref{eq:aic} and BIC \eqref{eq:bic} shows that BIC penalize the number of free parameter much more than AIC does. Therefore, it is generally believed that AIC tends to overfitting while BIC tends to underfitting \cite{10}. 

The existing regularized GLMs have their limitations for variable selection as well. As pointed out in \cite{elasticnet}, Ridge Regression cannot produce parsimonious models. When there are a group of correlated predictors, LASSO tends to select only one from the group \cite{elasticnet}, and to include many false positive variable into models while Elastic Net tends to select the variable group in or out together \cite{fan2010}. The oracle property of Adaptive LASSO is only attached to GLMs under certain restrictive regularity condition.

The ultimate evaluation criterion for variable selection methods is whether they can discover the ‘right’ variables for the response and hence derive an interpretable model for the given problem with good prediction performance. In many domains, the interpretablity of models is a much desired merit than the predictability of models. In this sense, the above existing methods cannot meet the requirement. 

Copula Entropy (CE) is a recently introduced multivariate statistical independence measure \cite{11} (more details in Section \ref{s:CopEnt}). It is defined rigorously with copula function and was proved to be equivalent to MI. However, it has not been applied to variable selection problem.

In Section \ref{subsec:measure}, two measures for statistical independence (dCor and dHSIC) are introduced. CE, dCor and dHSIC are all promising tools for variable selection. Recently, B\"ottcher defined copula versions of dCor and dHSIC \cite{bottcher2020}. However, no empirical comparison between these three measures has been done yet.

In this paper, a variable selection method based on CE \cite{11}, which can select the `right' variables for explainable models without sacrificing prediction ability, is proposed. Due to CE, the proposed method is theoretically sound and computationally efficient. The proposed method is evaluated on a biomedical dataset, and compared with the existing methods on variable selection.

\section{Copula Entropy}
\label{s:CopEnt}
\subsection{Theory}
Copula theory is about the representation of multivariate dependence with copula function \cite{12,13}. At the core of copula theory is Sklar theorem \cite{14} which states that multivariate probability density function can be represented as a product of its marginals and copula density function which represents dependence structure among random variables. Such representation seperates dependence structure, i.e., copula function, with the properties of individual variables -- marginals, which make it possible to deal with dependence structure only regardless of joint distribution and marginal distribution. This section is to define an statistical independence measure with copula. For clarity, please refer to \cite{11} for notations.

With copula density, Copula Entropy is define as follows \cite{11}:
\begin{definition}[Copula Entropy]
	\label{d:ce}
	Let $\mathbf{X}$ be random variables with marginal distributions $\mathbf{u}$ and copula density $c(\mathbf{u})$. CE of $\mathbf{X}$ is defined as
	\begin{equation}
	H_c(\mathbf{X})=-\int_{\mathbf{u}}{c(\mathbf{u})\log{c(\mathbf{u})}}d\mathbf{u}.
	\end{equation}
\end{definition}

In information theory, MI and entropy are two different concepts \cite{15}. In \cite{11}, Ma and Sun proved that they are essentially same -- MI is also a kind of entropy, negative CE, which is stated as follows: 
\begin{theorem}
	\label{thm1}
	MI of random variables is equivalent to negative CE:
	\begin{equation}
	I(\mathbf{X})=-H_c(\mathbf{X}).
	\end{equation}
\end{theorem}
\noindent
The proof of Theorem \ref{thm1} is simple \cite{11}. There is also an instant corollary (Corollary \ref{c:ce}) on the relationship between information of joint probability density function, marginal density function and copula density function.
\begin{corollary}
	\label{c:ce}
	\begin{equation}
	H(\mathbf{X})=\sum_{i}{H(X_i)}+H_c(\mathbf{X}).
	\end{equation}
\end{corollary}
The above results cast insight into the relationship between entropy, MI, and copula through CE, and therefore build a bridge between information theory and copula theory. CE itself provides a mathematical theory of statistical independence measure.

\subsection{Estimation}
\label{s:est}
It has been widely considered that estimating MI is notoriously difficult. Under the blessing of Theorem \ref{thm1}, Ma and Sun \cite{11} proposed a simple and elegant non-parametric method for estimating CE (MI) from data which comprises of only two steps\footnote{The \textsf{R} package \textsf{copent} for estimating CE is available on CRAN and also on GitHub at \url{https://github.com/majianthu/copent}.}:
\begin{enumerate}
	\item Estimating Empirical Copula Density (ECD);
	\item Estimating CE.
\end{enumerate}

For Step 1, if given data samples $\{\mathbf{x}_1,\ldots,\mathbf{x}_T\}$ i.i.d. generated from random variables $\mathbf{X}=\{x_1,\ldots,x_N\}^T$, one can easily estimate ECD as follows:
\begin{equation}
F_i(x_i)=\frac{1}{T}\sum_{t=1}^{T}{\chi(\mathbf{x}_{t}^{i}\leq x_i)},
\end{equation}
where $i=1,\ldots,N$ and $\chi$ represents for indicator function. Let $\mathbf{u}=[F_1,\ldots,F_N]$, and then one can derives a new samples set $\{\mathbf{u}_1,\ldots,\mathbf{u}_T\}$ as data from ECD $c(\mathbf{u})$. In practice, Step 1 can be easily implemented non-parametrically with rank statistics.

Once ECD is estimated, Step 2 is essentially a problem of entropy estimation which has been contributed with many existing methods. Among them, the kNN method \cite{kraskov} was suggested in \cite{11}. With rank statistic and kNN methods, one can derive a non-parametric method of estimating CE, which can be applied to any situation without any assumptions on the underlying system.

\section{CE based Variable Selection}
In this section, we propose a new variable selection method based on CE. It is a new kind of association measure based method. The idea is simple: the CE between response and predictors are estimated from data, and then predictors are selected according to the value of CE. With the selected variables, a model for the prediction problem is built. In the method, CE is estimated nonparametrically with the method in Section \ref{s:est}.

Since CE has many advantages over traditional association measure CC, it is quite obvious that the new method is superior to CC based method. The new method closely related to previously proposed MI based method but CE has more clear mathematical meaning for all the multivariate cases and is estimated in a new non-parametric way which makes the method efficient, stable and universally applicable. Since CE is defined as a distribution-free measure, the proposed method based on it is therefore model-free.

Another advantage of the new method is that the variable such selected can be attached with biological and physical meaning since CE, as a type of entropy, measures not only statistical dependence between variable and response, but also information transmission or energy exchange in the underlying systems. If a group of variables are selected by the proposed method, it is supposed to correspond to physical, biological or social meanings in the given system. As contrast, traditional methods have no such merit.

\section{Experiments and Results}
\subsection{Data}
The heart disease dataset in the UCI machine learning repository \cite{17} is used in our experiments, which contains 4 databases about heart disease diagnosis collected from four locations. The dataset includes 920 samples totally, of which only 899 samples without missing values are used in the experiments. All the dataset have the same instance format with 76 raw attributes, of which the attribute `num' is the diagnosis of patients’ disease. In the past research, only 14 attributes are recommended by researchers for clinical use, as listed in Table \ref{tb:attr} \cite{nahar2013}.
\begin{table}
	\centering
	\caption{The recommanded attributes in the UCI heart disease dataset.}
	\vskip2mm
	\begin{tabular}{l||c|c|c|c|c|c|c}
		\toprule
		\textbf{ID}&3&4&9&10&12&16&19\\
		\hline
		\textbf{Name}&age&sex&cp&trestbps&chol&fbs&restecg\\
		\hline
		\textbf{ID}&32&38&40&41&44&51&58\\
		\hline
		\textbf{Name}&thalach&exang&oldpeak&slope&ca&thal&num\\
		\bottomrule
	\end{tabular}
	\label{tb:attr}
\end{table}

\subsection{Evaluation Criteria}
To evaluate the variable selection methods, there are two criteria: predictability and interpretability. In our experiments, on one side, we suggest to test the prediction accuracy of the prediction models building on the selected variables. On the other side, we will try to check the explainability of the selected variable with reference to the established domain knowledge.

\subsection{Experiments}
We conducted 10 experiments on heart disease dataset to compare CE based method with other related methods. The goal of the experiments is to predict the diagnosis from other variables. In all experiments, both training data and test data were the whole dataset since we only investigate the variable selection ability of the methods and do not want to verify the generalizability of the models. The first experiment provides the baseline, in which 13 recommended variables were used and a SVM \cite{svm} classifier was trained on these variables. In the following three experiment, a group of variables are selected with the CE, dCor, and dHSIC based method, and then a SVM classifier is trained on such selected variables. The next two experiment is on stepwise selection on GLM with AIC and BIC. Since the response (the `num' attribute of heart disease dataset) has 5 levels of value, the GLMs in all the experiments are set as Poisson Regression model with the `log' link function. Stepwise selection is set as a `backward' one under the guidance of AIC and BIC. The next 3 experiments are on regularized GLMs, including LASSO, Ridge Regression, and Elastic Net with $\lambda_1,\lambda_2=0.5$. For each regularized GLM, the best amount of shrinkage is determined with 10-fold cross validation. The last experiment is on Adaptive LASSO with 10-fold cross validation to test its oracle property. The \textsf{R} packages `\textsf{copent}' \cite{copent},  `\textsf{energy}' \cite{dcorvar}, `\textsf{dHSIC}' \cite{dhsicvar}, `\textsf{e1071}' \cite{libsvm}, `\textsf{glmnet}' \cite{18}, and `\textsf{parcor}' \cite{zou2006ada,kraemer2009} were used for CE, dCor, dHSIC, SVM, regularized GLMs and Adaptive LASSO in the experiments respectively. The default values of the parameters are used for the functions of CE, dCor, and dHSIC in the corresponding packages.

\subsection{Results}
The prediction accuracy of the models are listed in Table \ref{tb:acc1}. It can be learned that the SVM classifier with variables selected by CE presents the best result (762 out of 899), better (5 more correct prediction, more than 0.56\% improvement) than SVM with the recommended variables. The other two dependence measure based method present comparable results. The remaining methods show only moderate prediction accuracy. Clearly, CE base variable selection improves the performance of prediction.

\begin{table}
	\centering
	\caption{Prediction accuracy of the models.}
	\vskip2mm
	\begin{tabular}{l|c}
		\toprule
		\textbf{Model}&\textbf{Accuracy} (\%)\\
		\midrule
		SVM (recommended variables) &84.20\\
		SVM (CE)	&\textbf{84.76}\\
		SVM (dCor)	&82.76\\
		SVM (dHSIC)	&84.54\\
		Stepwise GLM (AIC)&51.8\\
		Stepwise GLM (BIC)&49.1\\
		LASSO&79.2\\
		Ridge Regression&63.0\\
		Elastic Net&75.9\\
		Adaptive LASSO&35.7\\
		\bottomrule
	\end{tabular}
	\label{tb:acc1}
\end{table}

\begin{figure}
	\centering
	\subfigure[LASSO]{\includegraphics[width=0.4\textwidth]{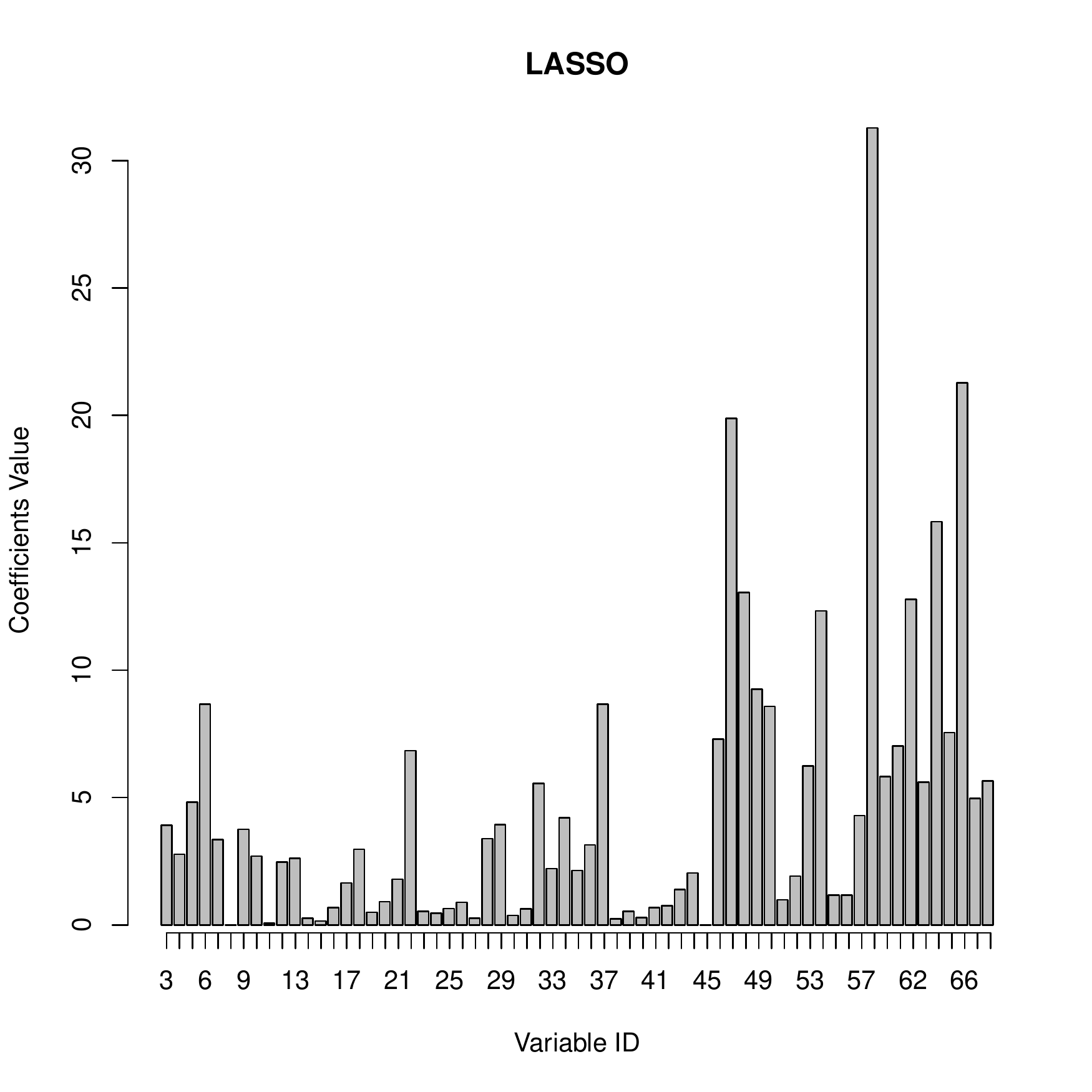}}	
	\subfigure[Ridge Regression]{\includegraphics[width=0.4\textwidth]{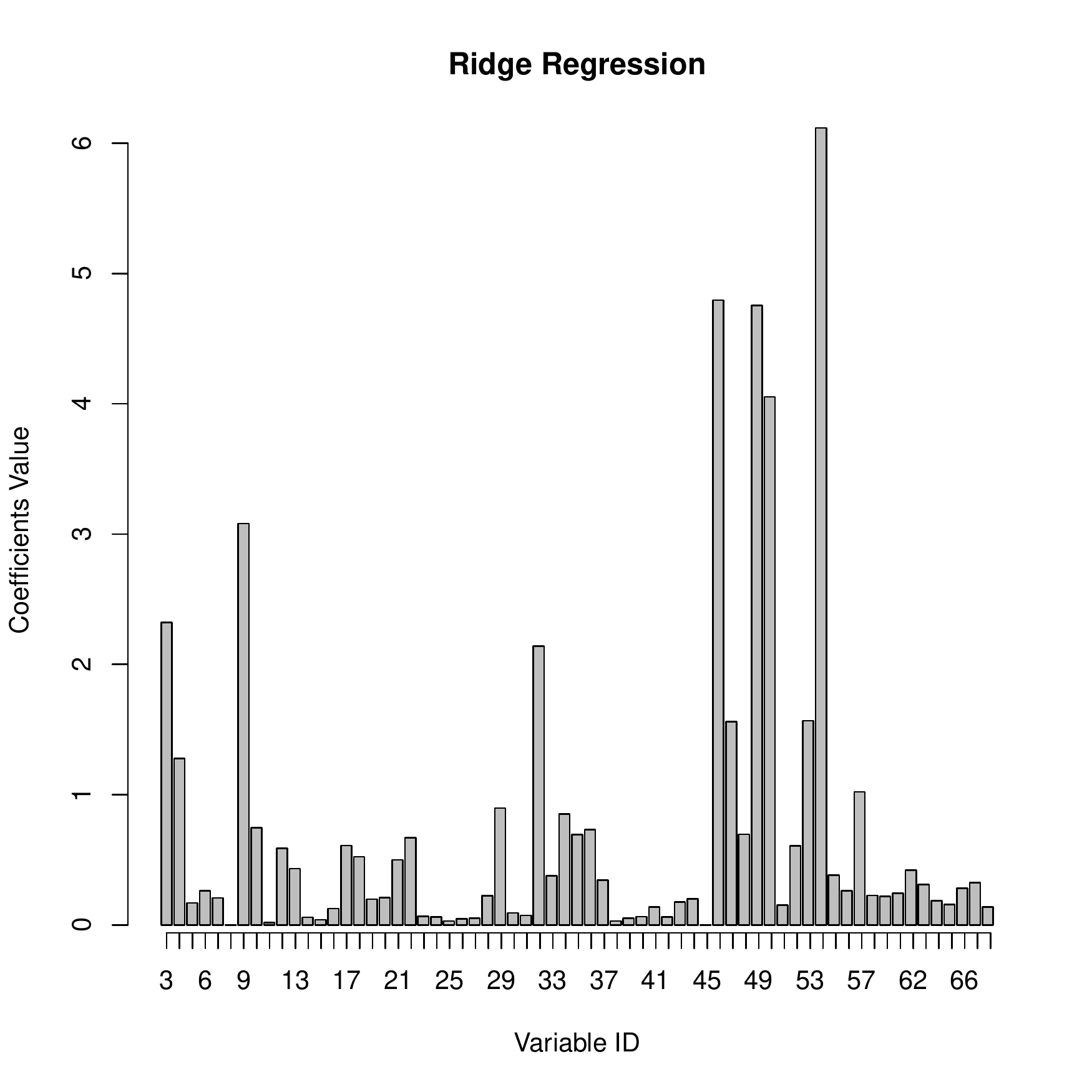}}
	\subfigure[Elastic Net]{\includegraphics[width=0.4\textwidth]{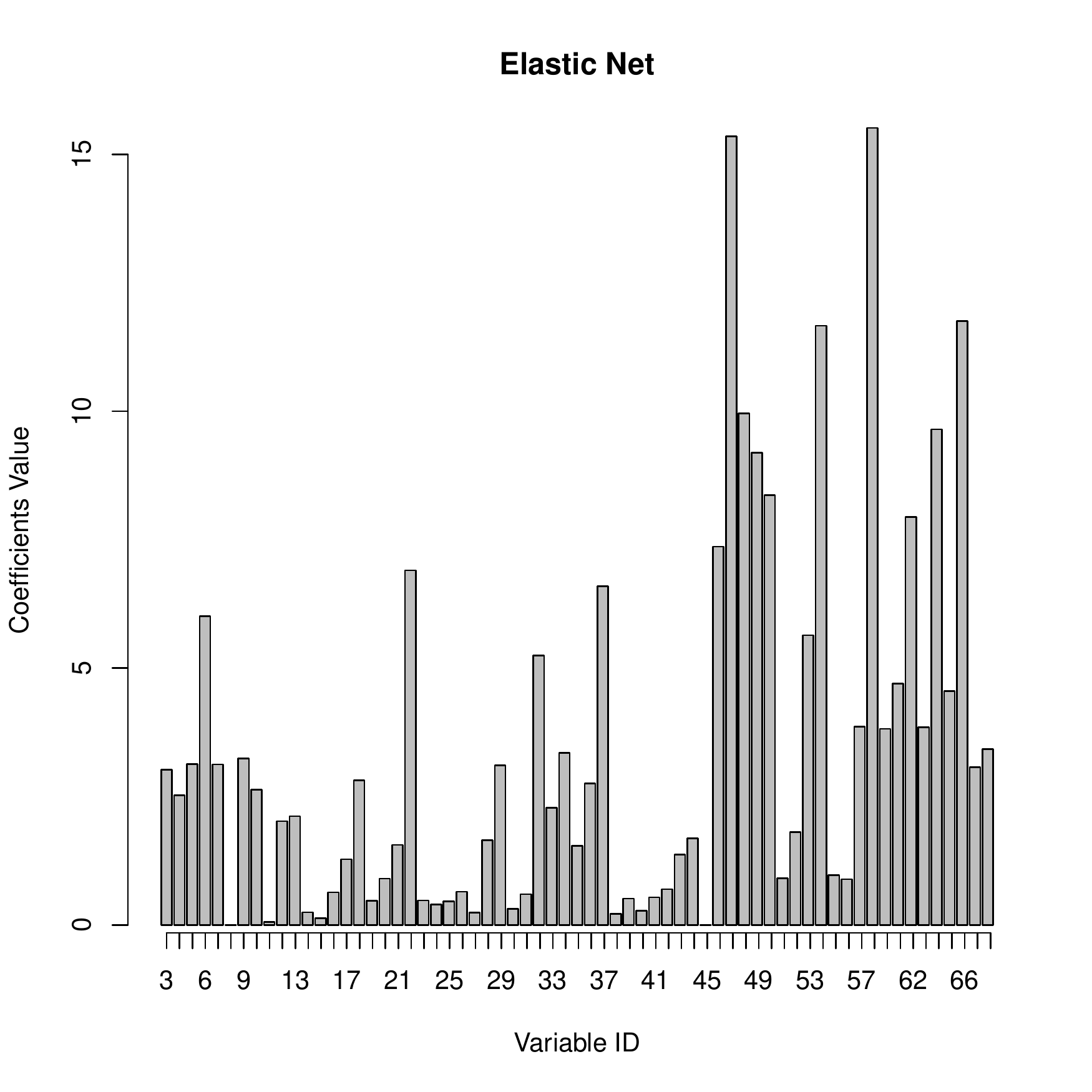}}	
	\subfigure[Adaptive LASSO]{\includegraphics[width=0.4\textwidth]{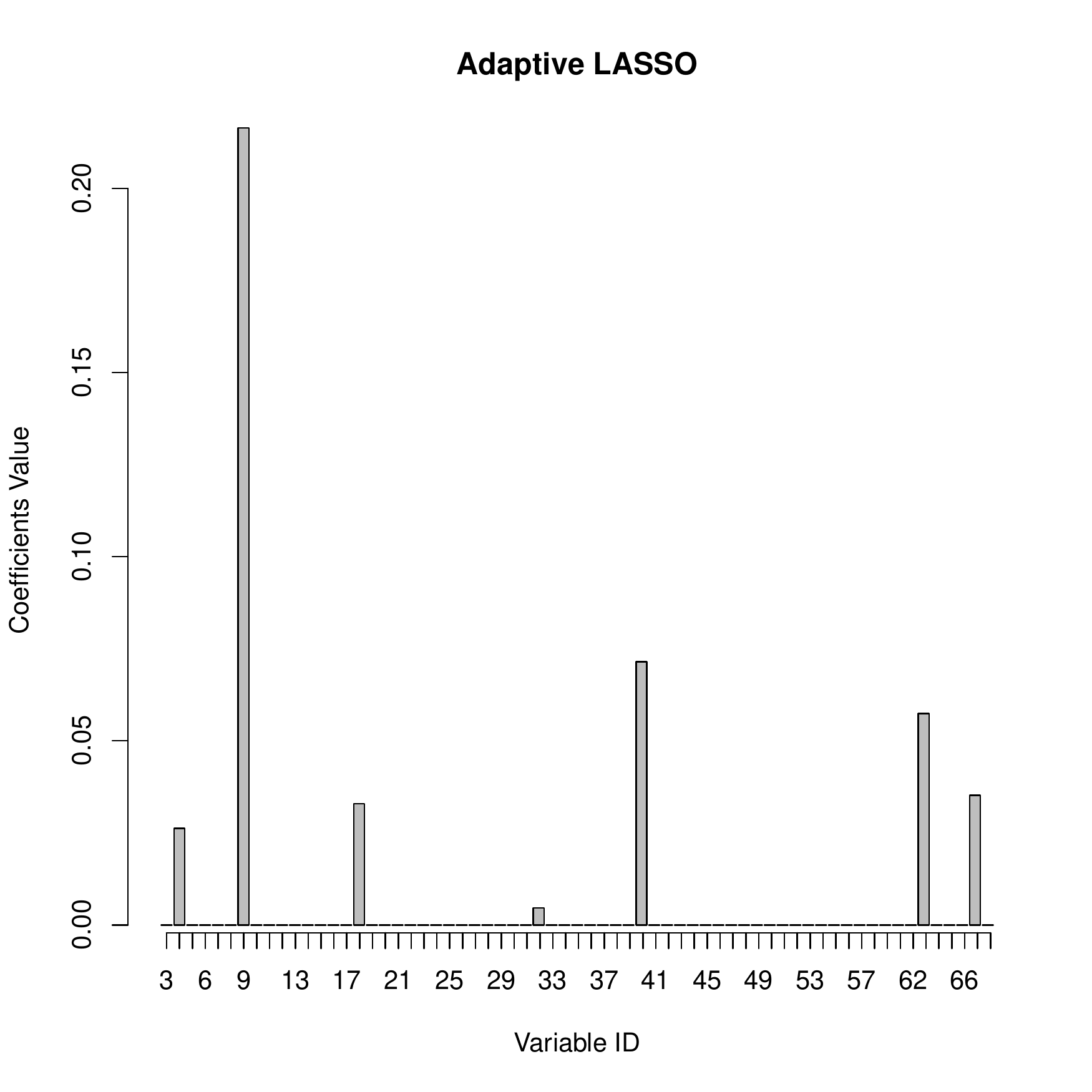}}	
	\subfigure[Stepwise GLM (AIC)]{\includegraphics[width=0.4\textwidth]{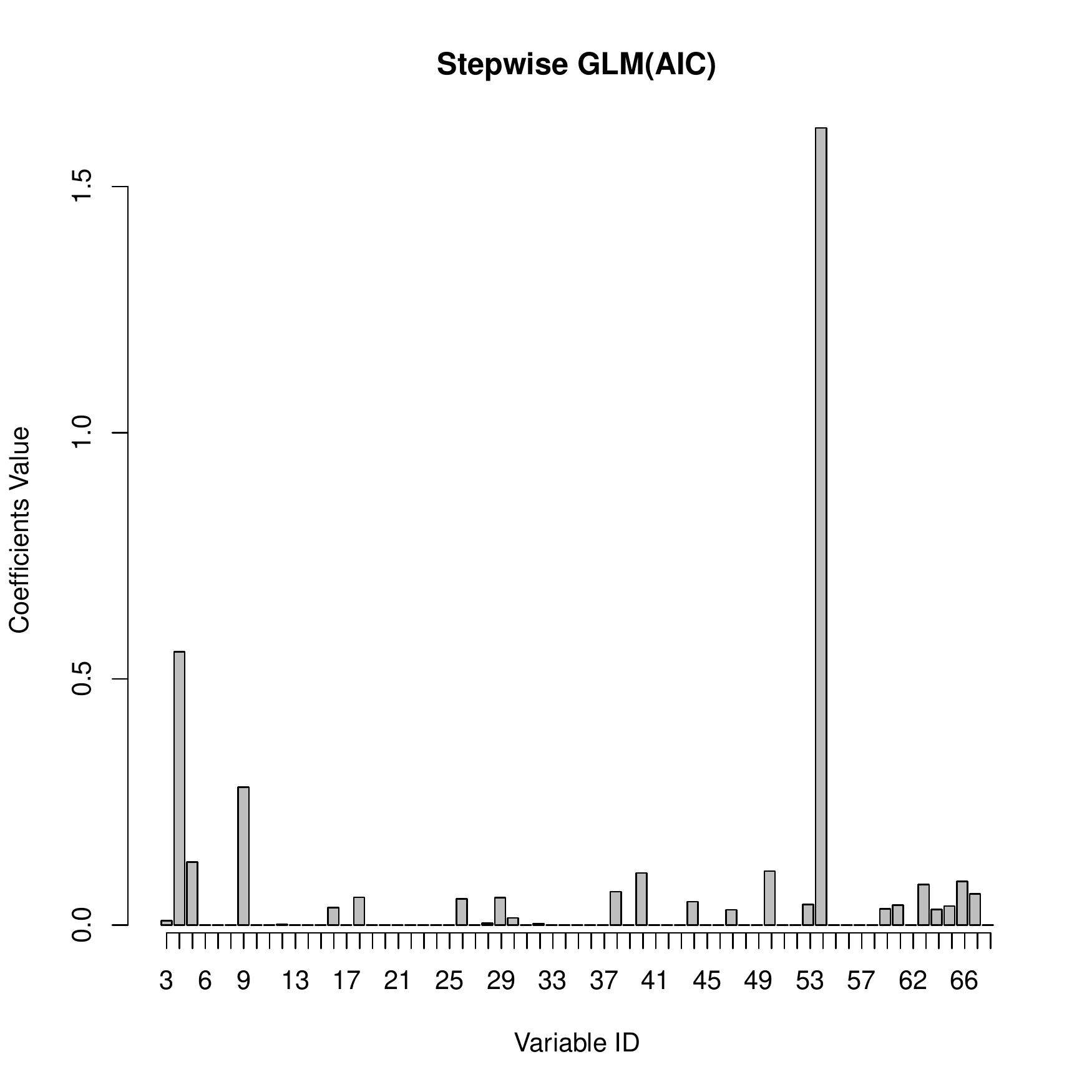}}	
	\subfigure[Stepwise GLM (BIC)]{\includegraphics[width=0.4\textwidth]{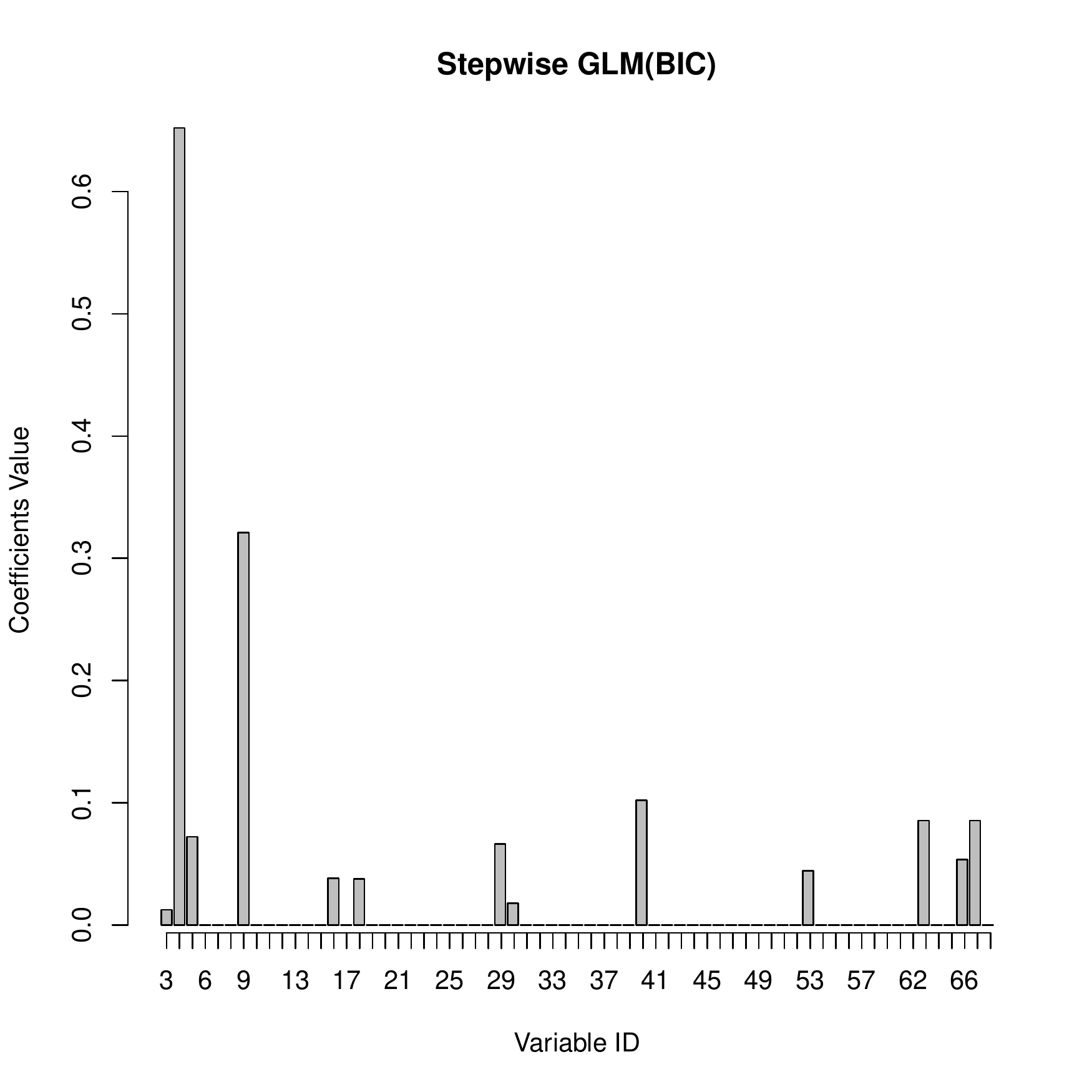}}	
	\caption{Coefficients of the GLMs.}
	\label{f:coef}
\end{figure}

Regularized GLMs select variables by non-zero coefficents. The coefficients of Regularized GLMs which are used in the prediction task are shown in Figure \ref{f:coef}. For multinomial logistic regression, there are 5 groups of coefficients for 5 levels of response. We take the mean of 5 groups of coefficients as the overall coefficients. They indicate the relative importance of the variables in each model. The variables corresponding to non-zero coefficients are considered as selected by the GLMs. 

\begin{figure}
	\centering
	\subfigure[CE]{\includegraphics[width=0.8\textwidth]{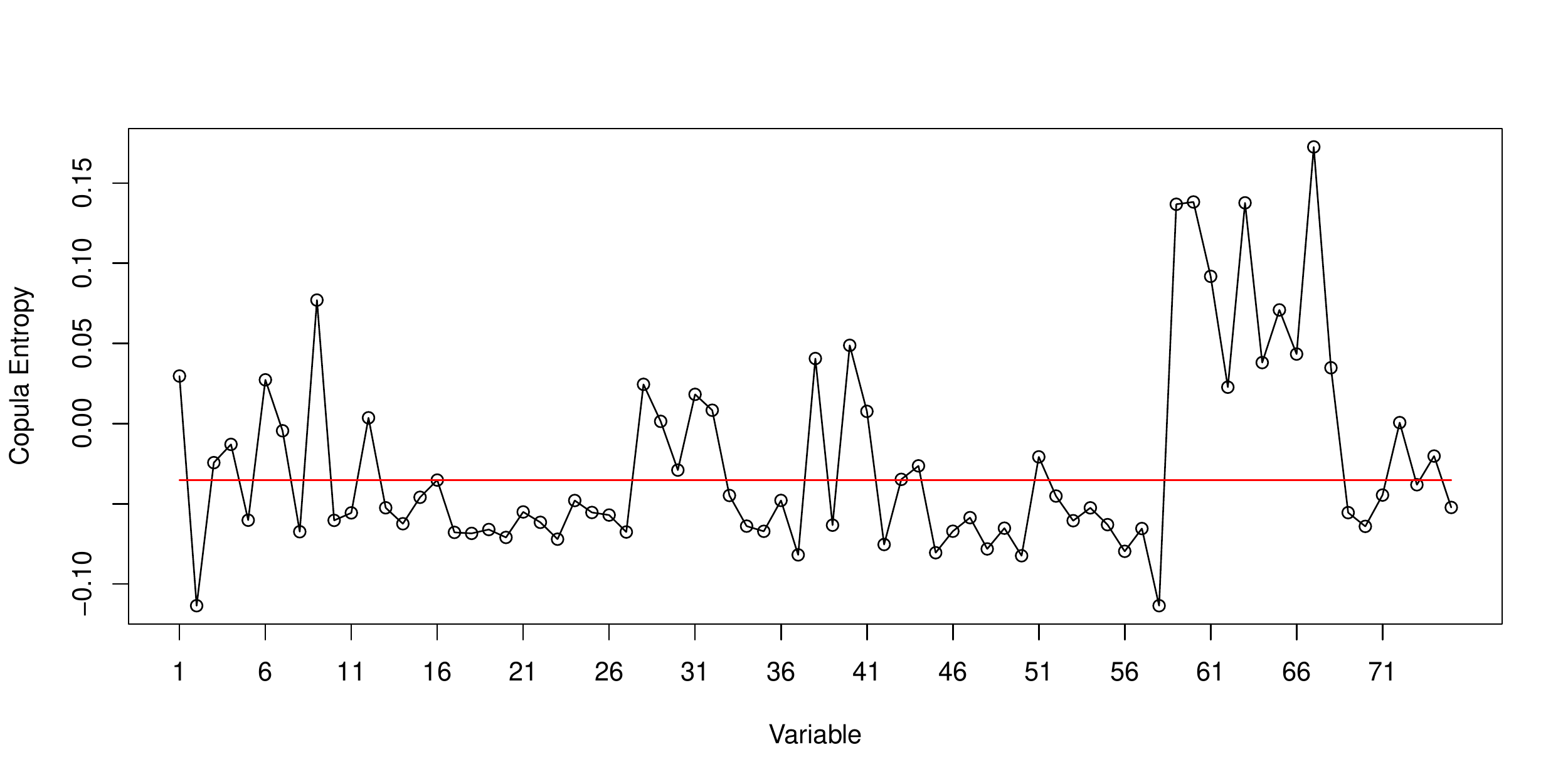}}	
	\subfigure[dCor]{\includegraphics[width=0.8\textwidth]{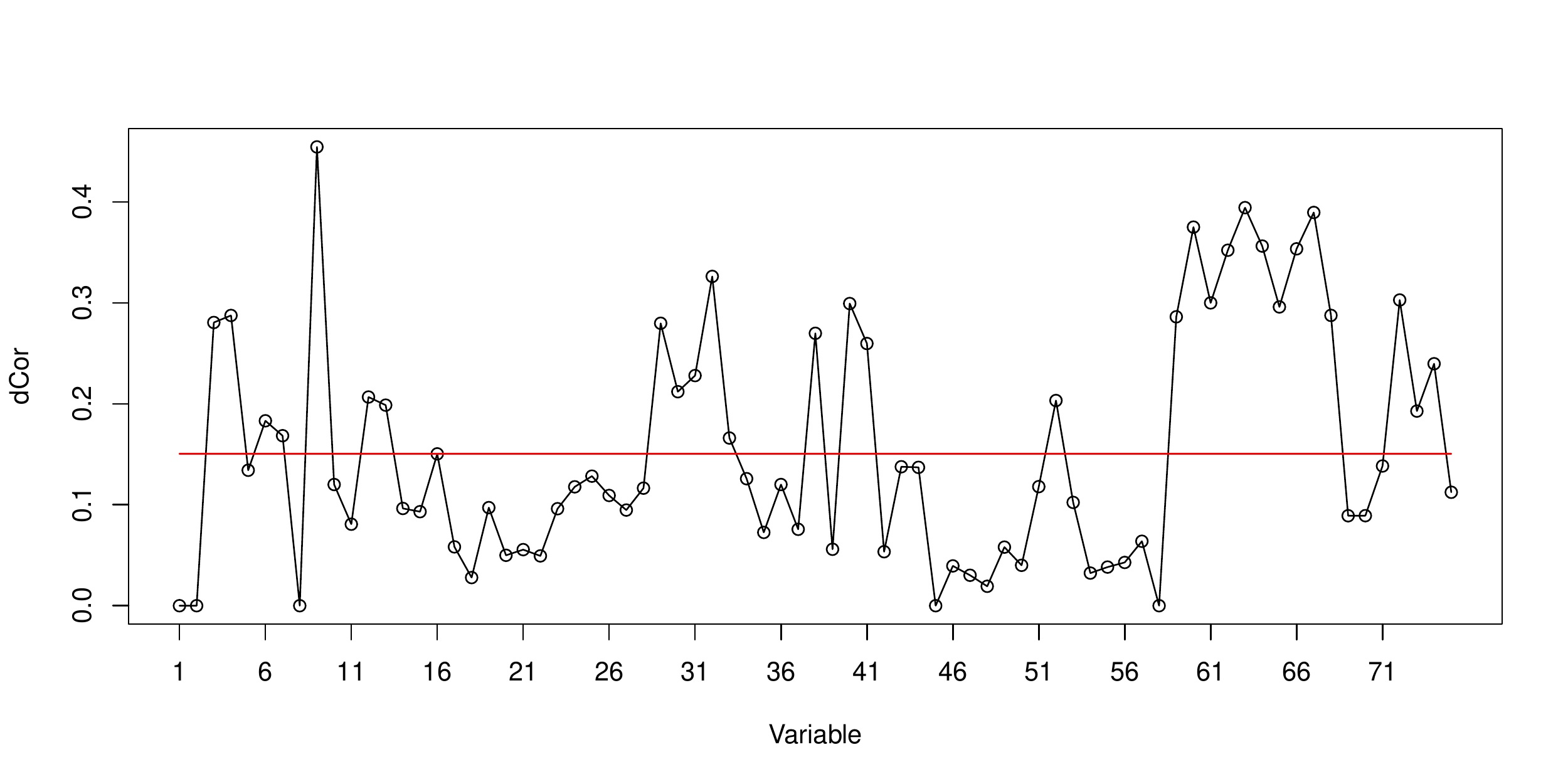}}	
	\subfigure[dHSIC]{\includegraphics[width=0.8\textwidth]{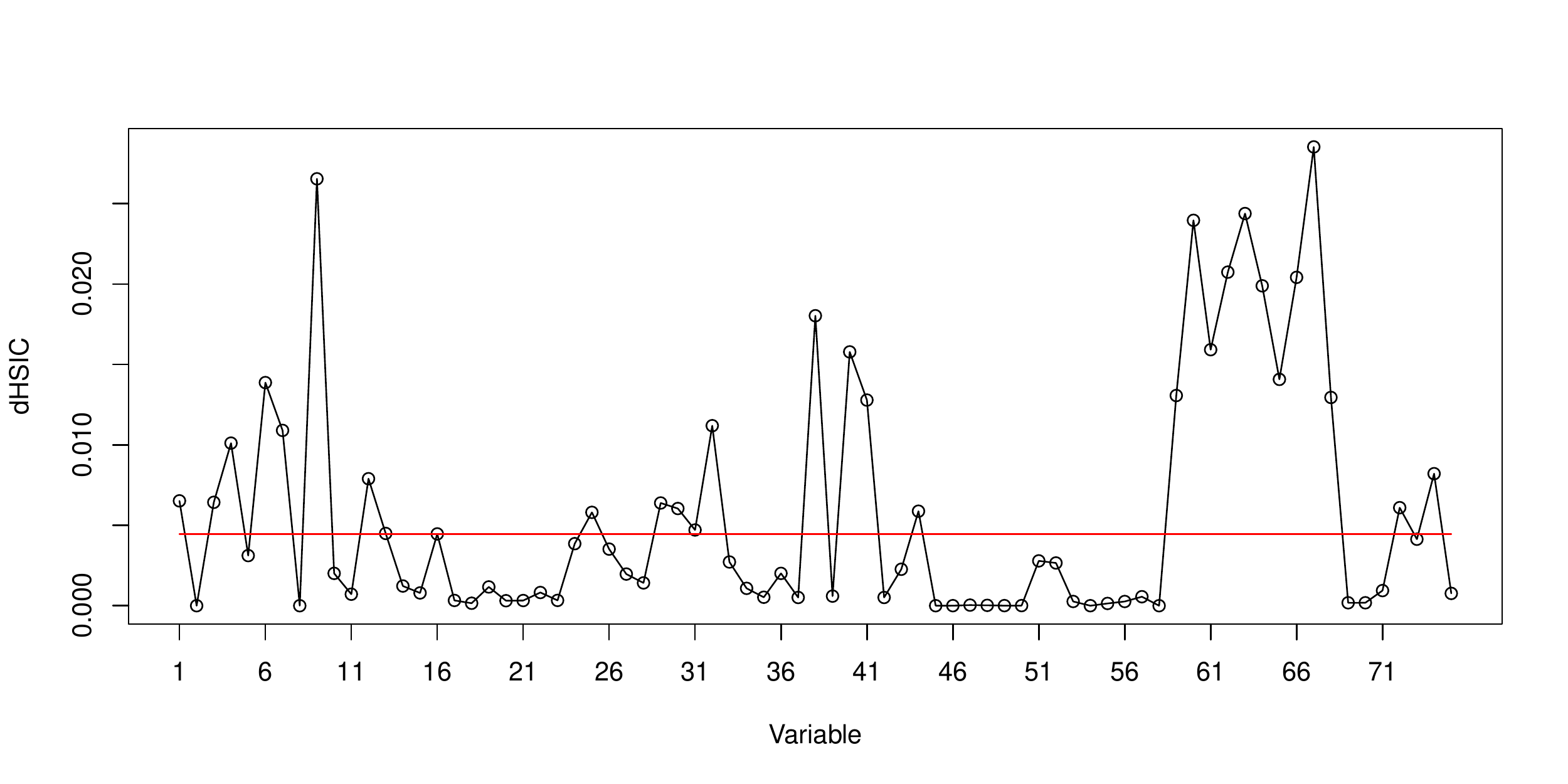}}	
	\caption{Variables selected by the three dependence measures.}
	\label{f:varsel}
\end{figure}

Variables selected by the three dependence measures based method are shown in Figure \ref{f:varsel}. The dependence strength of `fbs' (fasting blood sugar, \#16) is used as the thresold for selection of all the three methods. The variables selected based on ranks of CE, dCor, and dHSIC, are listed in Table \ref{tb:var1}. 

To check the interpretability of the selected variables of different methods, the recommended variables are taken as a golden rule for heart disease diagnosis since they are recommended by perfessional researchers as clinical relevant \cite{nahar2013}. The variables selected by different methods are summarized in Table \ref{tb:var1}. We compare the selected variables in each experiment with the recommended set. It can be learned from Table \ref{tb:var1} that CE based method selects 11 out of 13 recommended variables, dCor and dHSIC select 9 and 10 out of 13 variables respectively, and meanwhile CE selects a smaller variable set with fewer false positive variables than the other two measures do which means CE has higher selection accuracy. As contrast, experimental results show that regularized GLMs fail to select the recommended variables out of others and Adaptive LASSO selects 4 out of 13 variables. Meanwhile, Stepwise GLMs with AIC and BIC select 8 and 5 out of 13 variables respectively. It should be mentioned that and the variables (\#59-68) corresponding to the properties of vessels are also selected together, all with high CE value, which is meaningful and deserves further investigation in clinical practice. However, stepwise GLM does not select them all out. The results demonstate strong ability of CE based method on selecting meaningful variables against its compititors.

\begin{table}
	\centering
	\caption{Selected variables by different methods.}
	\vskip2mm
	\begin{tabular}{l|c}
		\toprule
		\textbf{Method} & \textbf{Variable ID}\\
		\midrule
		Recommandation&3,4,9,10,12,16,19,32,38,40,41,44,51\\
		\hline
		LASSO & all except 8,45\\
		\hline
		Ridge Regression & all except 8,45\\
		\hline
		Elastic Net & all except 8,45\\
		\hline
		Adaptive LASSO & 4,6,9,18,32,40,63,67\\
		\hline
		Stepwise GLM  (AIC)&\tabincell{c}{3,4,5,9,12,16,18,20,26,29,30,32,40,44,\\47,50,53,54,60,61,63,65-67}\\
		\hline
		Stepwise GLM (BIC)&3,4,5,9,16,18,29,30,40,53,63,66,67\\
		\hline
		dCor &3,4,6,7,9,12,13,16,28-33,38,40,41,52,59-68\\
		\hline
		dHSIC &3,4,6,7,9,12,13,16,25,29-32,38,40,41,44,59-68\\
		\hline
		CE	&3,4,6,7,9,12,16,28-32,38,40,41,44,51,59-68\\
		\bottomrule
	\end{tabular}
	\label{tb:var1}
\end{table}

\section{Discussion}
In the above experiments, the regularized GLMs select the variables based on the coefficients of the models. To achieve the goal, the methods should set up the models manually first, and then one has to tune both penalty parameter $\lambda_i$ and shrinkage amount during model training. This means making assumptions of the model of the underlying system, including sparsity of coefficients and specific nonlinearity, which are usually incorrect. Even though the free parameters has been tuned to optimal, experimental results show that regularized GLM failed to select the recommanded variables out and meanwhile they presented moderate prediction accuracy. Stepwise GLM do better than regularized GLMs on variable selection but presented poor prediction performance. 

Compared with them, CE based method is both model-free and tuning-free. It does nothing on model setting, tunes no parameter, and yet presents good performance on both prediction accuracy and variable selection. This is because that CE is a distribution-free measure of statistical indepedence and that its estimation is done in a non-parametric way. When applied, it makes no assumption on the underlying systems. With CE, variable selection is becoming a science with unversally applicable theory and efficient method, instead of an art like regularized GLMs and other information criteria.

In the experiments, dCor and dHSIC also presents comparable results and are also model-free and almost tuning-free. However, CE presents better results on both prediction accuracy and selection efficiency than dCor and dHSIC do. The good performance of CE over dCor and dHSIC can be explained theoretically. Though all the three measures characterize multivariate independence, CE has a much rigorous definition as a type of entropy which has many well-known axiomatic properties for statistical dependence. As contrast, dCor and dHSIC are essentially nonlinear generalizations of traditional correlation concept and do not has the axiomatic properties that CE has, such as invariant to monotonic transformation, equivalent to correlation coefficients in Gaussian cases. As a type of entropy, CE also enjoys intrinsic physical meaning as a measure of information or energy exchange in the underlying systems which makes the variables selected with CE interpretable while the physical meaning of dCor and dHSIC are unclear yet.

It can be learned that stepwise GLM selects out less recommended variables than the proposed method does. This is because that the selection criteria of the methods are different. AIC is essentially an approximation of KL divergence \cite{aic} between GLM and the underlying distribution. Since it is under the risks of model misspecification and approximation bias, the result is not so good as CE's. BIC is a criterion which is derived under the Bayesian framework with assumptions on the underlying models and the approximation of bayesian compuation \cite{bic}. It is usually unrealistic for BIC to assume that the true model is within the model family under consideration. As contrast to AIC and BIC, the proposed method is guided by the CE, a well defined and estimated measure instead of a divergence or posteriori, which makes no assumption on the underlying models so there are no above risks of model misspecification or unrealistic assumptions and hence the selection is more advantagous theorectially. Meanwhile, the proposed method is computationally effective and efficient while both AIC and BIC are kind of approximations computed under certain assumptions. 

The Adaptive LASSO which claims to has oracle property selects only a few `oracle' variable out. It is because that the Adaptive LASSO possesses it oracle property under restrictive model assumptions and regularity conditions. Comparing with it, CE presents very good results because the oracle property of CE is guaranteed by statistical dependence between variables and response measured by CE, which is unconditioned and universally applicable. 

CE also leads to interpretability of models. CE based variable selection is based on the dependence relationships between variables and response of model. It selects variables based on the dependence strength measured by CE, which is independent of variable's scale and dimension. Such statistical relationships are believed to have real world physical or biological meanings, as has been demonstrated in discovering statistical associations \cite{ma2019} -- a problem closely related to variable selection. The ability of the proposed method to select meaningful variables is also demonstrated in the above experiments where the variables group selected by CE is very similar to the variable group chosen by the perfessionals for clinical use and the additional vessels-related variables which are also meaningful but not considered before are also selected. The models built from the variables such selected can be explained with domain knowledge and applied to the cases where explainability matters. 

\section{Conclusion}
In this paper, we propose a CE based method for variable selection which use CE based ranks to select variables. The proposed method is both model-free and tuning-free. Comparison experiments between the CE based method and traditional variable selection methods, such as dCor, dHSIC, Stepwise Selection with information criteria, Regularized GLMs and Adaptive LASSO, are conducted on the UCI heart disease data. Experimental results show that CE based method can select the `right' variables out effectively and derive better interpretable results than traditional methods do without sacrificing predictability. It is believed that CE based method makes variable selection becoming a science instead of an art and can help to build more explainable models which can lead to successful applications where interpretability of model matters. In the future, the proposed method will be applied to more real-world dataset to test its effectiveness and efficiency.

\section*{Acknowledgement}
\noindent
The author thanks Matsumori Masaki for comments and suggestions.

\vspace{6mm}

\end{document}